\documentclass[11pt]{article}

\usepackage[utf8]{inputenc}
\usepackage[english]{babel}

\usepackage{hyperref}
\usepackage{wrapfig}
\usepackage[round]{natbib}

\usepackage{algorithm}
\usepackage{algpseudocode}
\algnewcommand\algorithmicforeach{\textbf{foreach}}
\algdef{S}[FOR]{ForEach}[1]{\algorithmicforeach\ #1\ \algorithmicdo}

\usepackage{mathtools}
\usepackage{amssymb, amsmath, amsthm}
\usepackage{xspace, float, graphicx}
\usepackage{mathabx}

\usepackage{caption}
\makeatletter
\@addtoreset{algorithm}{section}
\makeatother

\usepackage{subcaption}

\usepackage{listings}
\usepackage{color}
\definecolor{deepblue}{rgb}{0,0,0.5}
\definecolor{deepred}{rgb}{0.6,0,0}
\definecolor{deepgreen}{rgb}{0,0.5,0}
\definecolor{backcolour}{rgb}{0.95,0.95,0.92}
\definecolor{codegray}{rgb}{0.5,0.5,0.5}

\usepackage{accsupp}    
\newcommand{\noncopynumber}[1]{
	\BeginAccSupp{method=escape,ActualText={}}
	#1
	\EndAccSupp{}
}
\lstdefinestyle{Python}{
	language        = Python,
	backgroundcolor=\color{backcolour},
	basicstyle      = \ttfamily,
	keywordstyle    = \color{deepblue},
	stringstyle     = \color{deepgreen},
	commentstyle    = \color{codegray}\ttfamily,
	numberstyle=\tiny\color{codegray}\noncopynumber,
	columns=flexible,
	numbers=left,
	stepnumber=1
}

\lstdefinestyle{Mona}{
	basicstyle      = \ttfamily,
	numberstyle=\tiny\color{codegray}\noncopynumber,	
	columns=flexible,
	numbers=left,
	stepnumber=1	
}

\lstdefinelanguage{PDDL}{
  basicstyle      = \ttfamily,
  sensitive=false,    
  morecomment=[l]{;}, 
  alsoletter={:,-},   
  numberstyle=\tiny\color{codegray}\noncopynumber,	
  numbers=left,
  stepnumber=1,
  morekeywords={
    define,domain,problem,not,and,or,when,forall,exists,either,
    :domain,:requirements,:types,:objects,:constants,
    :predicates,:action,:parameters,:precondition,:effect,
    :fluents,:primary-effect,:side-effect,:init,:goal,
    :strips,:adl,:equality,:typing,:conditional-effects,
    :negative-preconditions,:disjunctive-preconditions,
    :existential-preconditions,:universal-preconditions,:quantified-preconditions,
    :functions,assign,increase,decrease,scale-up,scale-down,
    :metric,minimize,maximize,
    :durative-actions,:duration-inequalities,:continuous-effects,
    :durative-action,:duration,:condition
  }
}

\theoremstyle{plain}
\newtheorem{theorem}{Theorem}[section] 

\theoremstyle{definition}
\newtheorem{definition}{Definition}[section]
\newtheorem{example}{Example}[section]

\usepackage{titling}
\newcommand{\subtitle}[1]{%
  \posttitle{%
    \par\end{center}
    \begin{center}\large#1\end{center}
    \vskip0.5em}%
}

\newcommand{\D}{\mathcal{D}}
 \newcommand{\F}{\mathcal{F}}
\newcommand{\G}{\mathcal{G}}

\renewcommand{\O}{\mathcal{O}} \renewcommand{\P}{\mathcal{P}}

 \newcommand{\V}{\mathcal{V}}

\newcommand{\cla}[1]{\makebox[0pt]{\hss#1\hss}}
\newcommand{\Once}{%
  \sbox0{$\Diamond$}%
  \usebox0\kern-.5\wd0\cla{\raisebox{.1ex}{\scalebox{.7}[1]{$-$}}}\kern.5\wd0%
}

\newcommand{\LTL}{{\sc ltl}\xspace}
\newcommand{\LTLf}{{\sc ltl}$_f$\xspace}
\newcommand{\PLTLf}{{\sc pltl}$_f$\xspace}

\newcommand{\PDDL}{{\sc pddl}\xspace}
\newcommand{\FOND}{{\sc fond}\xspace}

\newcommand{\FONDS}{{\sc fond}$_{sp}$\xspace}
\newcommand{\FONDFOR}{{\sc fond}$4$\LTLf/\PLTLf}

\newcommand{\DFA}{{\sc dfa}\xspace}
\newcommand{\DFAs}{{\sc dfa}s\xspace}

\newcommand{\tup}[1]{\langle #1 \rangle}

\newcommand{\EXPTIME}{{\sc exptime}\xspace}

\newcommand{\TWOEXPTIME}{{\sc 2exptime}\xspace}

\newcommand{\trace}{\pi}

\newcommand{\automaton}{\mathcal{A}}

\newcommand{\lAND}{\wedge}

\newcommand{\LTLfToDFA}{{\sc ltl}$_f2$\DFA}

\algnewcommand\algInput{\textbf{input}}
\algnewcommand\algOutput{\textbf{output}}

\title{FOND Planning for \LTLf and \PLTLf Goals}
\subtitle{Extract of MSc. Thesis}
\author{Francesco Fuggitti}
\date{}

\begin{document}

\maketitle


\section*{Disclaimer for the Reader}
This is an extract of my Master thesis titled ``\LTL and Past \LTL on Finite Traces for Planning and Declarative Process Mining". The thesis was carried out in 2018 at Sapienza University under the supervision of Prof. Giuseppe De Giacomo. Since then, some tools developed in the thesis have been updated to major releases. Therefore, some of statements present in this extract may be referring to older releases of such tools. 

\section{Introduction}\label{ch:planning}
In this report, we will define a new approach to the problem of non deterministic planning for extended temporal goals. In particular, we will give a solution to this problem reducing it to a \emph{fully observable non deterministic} (\FOND) planning problem and taking advantage of the \LTLfToDFA\footnote{\href{http://ltlf2dfa.diag.uniroma1.it/}{http://ltlf2dfa.diag.uniroma1.it/}} tool. First of all, we will introduce the main idea and motivations supporting our approach. Then, we will give some preliminaries explaining the Planning Domain Definition Language (\PDDL) language and the \FOND planning problem formally. After that, we will illustrate our \FONDFOR (available online at \href{http://fond4ltlfpltl.diag.uniroma1.it/}{http://fond4ltlfpltl.diag.uniroma1.it/}) approach with the encoding of temporal goals into a \PDDL domain and problem. 
Finally, we will present some of the results obtained through the application of the proposed solution.

\section{Idea and Motivations}\label{sec:plan-idea-motiv}
Planning for temporally extended goals with \textit{deterministic} actions has been well studied during the years starting from \citep{bacchus1998planning} and \citep{doherty2001talplanner}. Two main reasons why temporally extended goals have been considered over the classical goals, viewed as a desirable set of final states to be reached, are because they are not limited in what they can specify and they allows us to restrict the manner used by the plan to reach the goals. Indeed, temporal extended goals are fundamental for the specification of a collection of real-world planning problems. Yet, many of these real-world planning problems have a \textit{non-deterministic} behavior owing to unpredictable environmental conditions. However, planning for temporally extended goals with \textit{non-deterministic} actions is a more challenging problem and has been of increasingly interest only in recent years with \citep{camacho2017non, de2018automata}.

In this scenario, we have devised a solution to this problem that exploits the translation of a temporal formula to a \DFA, using \LTLfToDFA. In particular, our idea is the following: given a non-deterministic planning problem and a temporal formula, we first obtain the corresponding \DFA of the temporal formula through \LTLfToDFA, then, we encode such a \DFA into the non-deterministic planning domain. As a result, we have reduced the original problem to a classic \FOND planning problem. In other words, we compile extended temporal goals together with the original planning domain, specified in \PDDL, which is suitable for input to standard (\FOND) planners.
\section{Preliminaries}
In this section, we will give some basics on the \PDDL specification language for domains and problems of planning and a general formalization of \FOND planning.
\subsection{\PDDL}\label{sec:pddl}
As stated before, \PDDL is the acronym for Planning Domain Definition Language, which is the \textit{de-facto} standard language for representing ``classical'' planning tasks. A general planning task has the following components:
\begin{itemize}
\item Objects: elements in the world that are of our interest;
\item Predicates: objects properties that can be true or false;
\item Initial state: state of the world where we start;
\item Goal state: things we want to be true;
\item Action/Operator: rule that changes the state. 
\end{itemize}
Moreover, planning tasks are composed by two files: the \textit{domain} file where are defined predicates and actions and a \textit{problem} file where are defined objects, the initial state and the goal specification.
\subsubsection{The \textit{domain} file} 
The \textit{domain} definition gives each domain a name and specifies predicates and actions available in the domain. It might also specify types, constants and other things. A simple domain has the following format:
\begin{lstlisting}[language=PDDL, escapechar=£, label={code:pddl-domain}]
(define (domain DOMAIN_NAME)
  (:requirements [:strips] [:equality] [:typing] [:adl] ...)
  [(:types T1 T2 T3 T4 ...)]
  (:predicates (PREDICATE_1_NAME [?A1 ?A2 ... ?AN])
                (PREDICATE_2_NAME [?A1 ?A2 ... ?AN])
	       ...)

  (:action ACTION_1_NAME
    [:parameters (?P1 ?P2 ... ?PN)]
    [:precondition PRECOND_FORMULA]
    [:effect EFFECT_FORMULA]
   )
  (:action ACTION_2_NAME
    ...)
  ...)  
\end{lstlisting}
where \texttt{[]} indicates optional elements. To begin with, any \PDDL \textit{domain} definition must declare its expressivity requirements given after the \texttt{:requirements} key. The basic \PDDL expressivity is called \textsc{strips}\footnote{\textsc{strips} stands for STanford Research Institute Problem Solver, which is a formal language of inputs to the homonym automated planner developed in 1971.}, whereas a more complex one is the Action Description Language (\textsc{adl}), that extends \textsc{strips} in several ways, such as providing support for negative preconditions, disjunctive preconditions, quantifiers, conditional effects etc.. Nevertheless, many planners do not support full \textsc{adl} because creating plans efficiently is not trivial. Although the presence of this limitation, the \PDDL language allows us to use only some of the \textsc{adl} features. Furthermore, there are also other requirements often used that can be specified as \texttt{equality}, allowing the usage of the predicate \texttt{=} interpreted as equality, and \texttt{typing} allowing the typing of objects. As we will explain later, our practical implementation supports, so far, only simple \textsc{adl}, namely conditional effects in domain's operators which do not have any nested subformula.

Secondly, there is the predicates definition after the \texttt{:predicates} key. Predicates may have zero or more parameters variables and they specify only the number of arguments that a predicate should have. Moreover, a predicate may also have typed parameters written as \texttt{?X -- TYPE\_OF\_X}.

Thirdly, there is a list of action definitions. An action is composed by the following items:
\begin{itemize}
\item \textit{parameters}: they stand for free variables and are represented with a preceding question mark \texttt{?};
\item \textit{precondition}: it tells when an action can be applied and, depending on given requirements, it could be differently defined (i.e. conjunctive formula, disjunctive formula, quantified formula, etc.);
\item \textit{effect}: it tells what changes in the state after having applied the action. As for the precondition, depending on given requirements, it could be differently defined (i.e. conjunctive formula, conditional formula, universally quantified formula, etc.)
\end{itemize}

In particular, in pure \textsc{strips} domains, the precondition formula can be one of the following:
\begin{itemize}
\item an atomic formula as \texttt{(PREDICATE\_NAME ARG1 ... ARG\_N)}
\item a conjunction of atomic formulas as \texttt{(and ATOM1 ... ATOM\_N)}
\end{itemize}
where arguments must either be parameters of the action or constants.

If the \textit{domain} uses the \texttt{:adl} or \texttt{:negated-precondition} an atomic formula could be expressed also as  \texttt{(not (PREDICATE\_NAME ARG1 ... ARG\_N))}. In addition, if the domain uses \texttt{:equality}, an atomic formula may also be of the form \texttt{(= ARG1 ARG2)}.

On the contrary, in \textsc{adl} domains, a precondition formula could be one of the following:
\begin{itemize}
\item a general negation as \texttt{(not CONDITION\_FORMULA)}
\item a conjunction of condition formulas as \texttt{(and CONDITION\_FORMULA1 ... \\CONDITION\_FORMULA\_N)}
\item a disjunction of condition formulas as \texttt{(or CONDITION\_FORMULA1 ... \\CONDITION\_FORMULA\_N)}
\item an implication as \texttt{(imply CONDITION\_FORMULA1 ... CONDITION\_FORMULA\_N)}
\item an implication as \texttt{(imply CONDITION\_FORMULA1 ... CONDITION\_FORMULA\_N)}
\item a universally quantified formula as \texttt{(forall (?V1 ?V2 ...) CONDITION\_FORMULA)}
\item an existentially quantified formula as \texttt{(exists (?V1 ?V2 ...)\\ CONDITION\_FORMULA)}
\end{itemize}

The same division can be carried out with effects formulas. Specifically, in pure \textsc{strips} domains, the precondition formula can be one of the following:
\begin{itemize}
\item an added atom as \texttt{(PREDICATE\_NAME ARG1 ... ARG\_N)}
\item a deleted atom as \texttt{(not (PREDICATE\_NAME ARG1 ... ARG\_N))}
\item a conjunction of effects as \texttt{(and ATOM1 ... ATOM\_N)}
\end{itemize}
On the other hand, in an \textsc{adl} domains, an effect formula can be expressed as:
\begin{itemize}
\item a conditional effect as \texttt{(when CONDITION\_FORMULA EFFECT\_FORMULA)}, where the \texttt{EFFECT\_FORMULA} is occur only if the \texttt{CONDITION\_FORMULA} holds true. A conditional effect can be placed within quantification formulas.
\item a universally quantified formula as \texttt{(forall (?V1 ?V2 ...) EFFECT\_FORMULA)}
\end{itemize}

As last remark that we will deepen later in Section \ref{sec:fond}, when the \PDDL \textit{domain} has \textit{non-deterministic} actions, the effect formula of those actions expresses the non-determinism with the keyword \texttt{oneof} as \texttt{(oneof (EFFECT\_FORMULA\_1) ... \\ (EFFECT\_FORMULA\_N)}.

In the following, we show a simple example of \PDDL \textit{domain}.

\begin{example}\label{ex:pddl-domain}
A simple \PDDL \textit{domain} the Tower of Hanoi game. This game consists of three rods and $n$ disks of different size, which can slide into any rods. At the beginning, disks are arranged in a neat stack in ascending order of size on a rod, the smallest on the top. The goal of the game is to move the whole stack to another rod, following three rules:
\begin{itemize}
\item one disk at a time can be moved;
\item a disk can be moved only if it is the uppermost disk on a stack;
\item no disk can be placed on top of a smaller disk.
\end{itemize}
\begin{lstlisting}[language=PDDL, escapechar=£]
(define (domain hanoi) ;comment£\label{line:domain-name}£
  (:requirements :strips :negative-preconditions :equality)£\label{line:requirements}£
  (:predicates (clear ?x) (on ?x ?y) (smaller ?x ?y) )£\label{line:predicates}£
  (:action move
    :parameters (?disc ?from ?to)£\label{line:parameters}£
    :precondition (and£\label{line:precond}£
       (smaller ?disc ?to) (smaller ?disc ?from)
       (on ?disc ?from)
       (clear ?disc) (clear ?to)
       (not (= ?from ?to))
    )
    :effect (and£\label{line:effects}£
      (clear ?from)
      (on ?disc ?to)
      (not (on ?disc ?from))
      (not (clear ?to))
    )
  )
)
\end{lstlisting}
The \PDDL \textit{domain} file of the Tower of Hanoi is quite simple. Indeed, it consists of only one action (\texttt{move}) and only a few predicates. Firstly, the name given to this \textit{domain} is \texttt{hanoi}. Then, there have been specified requirements as \texttt{:strips}, \texttt{:negative-preconditions} and \texttt{:equality}. After that, at line \ref{line:predicates}, there is the definition of all predicates involved in the \PDDL \textit{domain}. In particular, there are three predicates to describe if the top of a disk is \texttt{clear}, which disk is \texttt{on} top of another and, finally, which disk is \texttt{smaller} than another. Finally, there is the \texttt{move} action declaration with its parameters, its precondition formula and its effect formula.
\end{example}
\subsubsection{The \textit{problem} file} 
After having examined how a \PDDL \textit{domain} is defined, we can see the formulation of a \PDDL \textit{problem}. A \PDDL \textit{problem} is what a planner tries to solve. The \textit{problem} file has the following format:
\begin{lstlisting}[language=PDDL, escapechar=£, label={code:pddl-domain}]
(define (problem PROBLEM_NAME)
  (:domain DOMAIN_NAME)
  (:objects OBJ1 OBJ2 ... OBJ_N)
  (:init ATOM1 ATOM2 ... ATOM_N)
  (:goal CONDITION_FORMULA)
  )
\end{lstlisting}
At first glance, we can notice that the \textit{problem} definition includes the specification of the domain to which it is related. Indeed, every problem is defined with respect to a precise \textit{domain}. Then, there is the object list which could be typed or untyped. After that, there are the initial and goal specification, respectively. The former defines what is true at the beginning of the planning task and it consists of ground atoms, namely predicates instantiated with previously defined objects. Finally, the goal description represents the formula, consisting of instantiated predicates, that we would like to achieve and obtain as a final state.
In the following, we show a simple example of \PDDL \textit{problem}.

\begin{example}
In this example, we show a possible \PDDL \textit{problem} for the Tower of Hanoi game for which we have shown the \textit{domain} in the Example \ref{ex:pddl-domain}.
\begin{lstlisting}[language=PDDL, escapechar=£]
(define (problem hanoi-prob)
  (:domain hanoi)
  (:objects rod1 rod2 rod3 d1 d2 d3)£\label{line:objs-prob}£
  (:init 
     (smaller d1 rod1) (smaller d2 rod1) (smaller d3 rod1)
     (smaller d1 rod2) (smaller d2 rod2) (smaller d3 rod2)
     (smaller d1 rod3) (smaller d2 rod3) (smaller d3 rod3)
     (smaller d2 d1) (smaller d3 d1) (smaller d3 d2)
     (clear rod2) (clear rod3) (clear d1)
     (on d3 rod1) (on d2 d3) (on d1 d2))
  (:goal (and (on d3 rod3) (on d2 d3) (on d1 d2)))
)
\end{lstlisting}
At line \ref{line:objs-prob}, we have three rods and three disks. At the beginning, all instantiated predicates that are true are mentioned. If a predicate is not mentioned, it is considered to be false. In the initial situation there have been specified all possible movements with the \texttt{smaller} predicate, the disks are one on top of the other in ascending order on \texttt{rod1} whereas the other two rods are \texttt{clear}. In addition, the goal description is a conjuctive formula requiring disks on a stack on the \texttt{rod3}.
\end{example}

Once both \PDDL \textit{domain} and a \textit{problem} are specified, they are given as input to planners.
\subsection{Fully Observable Non Deterministic Planning}\label{sec:fond}
In this section, we formally define what \textit{Fully Observable Non Deterministic Planning} (\FOND) is giving some notions and definitions. Initially, we recall some concepts of ``classical'' planning while assuming the reader to be acquainted with basics of planning.

Given a \PDDL specification with a \textit{domain} and its corresponding \textit{problem}, we would like to solve this specification in order to find a sequence of actions such that the goal formula holds true at the end of the execution. A \textit{plan} is exactly that sequence of actions which leads the agent to achieve the goal starting from the initial state. Formally, we give the following definition.

\begin{definition}\label{def:classic-planning}
A planning problem is defined as a tuple $\P = \tup{\Sigma, s_0, g}$, where:
\begin{itemize}
\item $\Sigma$ is the state-transition system;
\item $s_0$ is the initial state;
\item $g$ is the goal state.
\end{itemize}
\end{definition}

\noindent Given the above Definition \ref{def:classic-planning}, we can formally define what a plan is.

\begin{definition}\label{def:plan}
A \textit{plan} is any sequence of actions $\sigma = \tup{a_1,a_2,\dots, a_n}$ such that each $a_i$ is a ground instance of an operator defined in the domain description.
\end{definition}

\noindent Moreover, we have that:

\begin{definition}\label{def:plan-sol}
A \textit{plan} is a solution for $\P = \tup{\Sigma, s_0, g}$, if it is executable and achieves $g$.
\end{definition}

Furthermore, a ``classical'' planning problem, just defined, is given under the assumptions of \textit{fully observability} and \textit{determinism}. In particular, the former means that the agent can always see the entire state of the environment whereas the latter means that the execution of an action is certain, namely any action that the agent takes uniquely determines its outcome.

Unlike the ``classical'' planning approach, in this thesis we focus on \textit{Fully Observable Non Deterministic} (\FOND) planning. Indeed, we continue relying on the \textit{fully observability}, but losing the \textit{determinism}. In other words, in \FOND planning we have the uncertainty on the outcome of an action execution. As anticipated in Section \ref{sec:pddl}, the uncertainty of the outcome of an operator execution is syntactically expressed, in \PDDL, with the keyword \texttt{oneof}. To better capture this concept, we give the following example.

\begin{example}
Here, we show as example the \texttt{put-on-block} operator of the \FOND version of the well-known blocksworld \PDDL \textit{domain}.
\begin{lstlisting}[language=PDDL, escapechar=£]
(:action put-on-block
  :parameters (?b1 ?b2 - block)
  :precondition (and (holding ?b1) (clear ?b2))
  :effect (oneof (and (on ?b1 ?b2) (emptyhand) (clear ?b1) £\label{line:oneof}£
                 (not (holding ?b1)) (not (clear ?b2)))
                 (and (on-table ?b1) (emptyhand) (clear ?b1) 
                 (not (holding ?b1)))
          )
)
\end{lstlisting}
The effect of the \texttt{put-on-block} is non deterministic. Specifically, the action is executed every time the agent is holding a block and another block is clear on the top. The effect can be either that the block is put on top of the other block or that the block is put on table. This has to be intended as an aleatory event. Indeed, the agent does not control the operator execution result.
\end{example}

Additionally, a \textit{non-deterministic} action $a$ with effect $oneof(E_1,\dots,E_n)$ can be intended as a set of \textit{deterministic} actions $\{b_1,\dots,b_n\}$, sharing the same precondition of $a$, but with effects $E_1,\dots,E_n$, respectively. Hence, the application of action $a$ turns out in the application of one of the actions $b_i$, chosen non-deterministically.

At this point, we can formally define the \FOND planning. Following \citep{ghallab2004automated} and \citep{geffner2013concise}, we give the following definition:
\begin{definition}
A \textit{non-deterministic domain} is a tuple $\D = \tup{2^\F, A, s_0, \varrho, \alpha}$ where:
\begin{itemize}
\item $\F$ is a set of \textit{fluents} (atomic propositions);
\item $A$ is a set of \textit{actions} (atomic symbols);
\item $2^\F$ is the set of states;
\item $s_0$ is the initial state (initial assignment to fluents);
\item $\alpha(s) \subseteq A$ represents \textit{action preconditions};
\item $(s, a, s') \in \varrho$ with $a \in \alpha(s)$ represents \textit{action effects} (including frame assumptions).
\end{itemize}
\end{definition}
\noindent Such domain $\D$ is assumed to be represented compactly (e.g. in \PDDL), therefore, considering the \textit{size} of the domain as the cardinality of $\F$. Intuitively, the evolution of a non-deterministic domain is as follows: from a given state $s$, the agent chooses what action $a \in \alpha(s)$ to execute, then, the environment chooses a \textit{successor state} $s'$ with $(s,a,s') \in \varrho$. To this extent, planning can also be seen as a \textit{game} between two players: the agent tries to force eventually reaching the goal no matter how the environment behaves.
Moreover, the agent can execute an action having the knowledge of all history of states so far.

Now, we can define the meaning of solving a \FOND planning problem on $\D$. A \textit{trace} of $\D$ is a finite or infinite sequence $s_0,a_0,s_1,a_1, \dots$ where $s_0$ is the initial state, $a_i \in \alpha(s_i)$ and $s_{i+1} \in \varrho(s_i,a_i)$ for each $s_i,a_i$ in the trace.

Solutions to a \FOND problem $\P$ are called \textit{strategies} (or \textit{policies}). A \textit{strategy} $\trace$ is defined as follows:
\begin{definition}\label{def:policy}
Given a \FOND problem $\P$, a \textit{strategy} $\trace$ for $\P$ is a partial function defined as:
\begin{equation}
\trace: (2^\F)^+ \rightarrow A
\end{equation}
such that for every $u \in (2^\F)^+$, if $\trace(u)$ is defined, then $\trace(u) \in \alpha(last(u))$, namely it selects applicable actions, whereas, if $\trace(u)$ is undefined, then $\trace(u) = \bot$.
\end{definition}
\noindent A trace $\tau$ is \textit{generated} by $\trace$ (often called $\trace$-trace) if the following holds:
\begin{itemize}
\item if $s_0,a_0,\dots,s_i,a_i$ is a prefix of $\tau$, then $\trace(s_0,s_1,\dots,s_i) = a_i$;
\item if $\tau$ is finite, i.e. $\tau = s_0,a_0,\dots,a_{n-1},s_n$, then $\trace(s_0,s_1,\dots,s_i) = \bot$.
\end{itemize}

For \FOND planning problems, in \citep{cimatti2003weak}, are defined different classes of solutions. Here we examine only two of them, namely \textit{strong solution} and \textit{strong cyclic solutions}. In the following, we give their formal definitions.

\begin{definition}\label{def:strong-sol}
A \textit{strong solution} is a strategy that is guaranteed to achieve the goal regardless of non-determinism.
\end{definition}

\begin{definition}\label{def:strong-cyc-sol}
\textit{Strong cyclic solutions} guarantee goal reachability only under the assumption of \textit{fairness}. In the presence of \textit{fairness} it is supposed that all action outcomes, in a given state, would occur infinitely often. 
\end{definition}
Obviously, \textit{strong cyclic solutions} are less restrictive than a \textit{strong solution}.
Indeed, as the name suggests, a strong cyclic solution may revisit states. However, in this thesis we will focus only on searching \textit{strong solutions}. As final remark, when searching for a strong solution to a \FOND problem we refer to \FONDS.

In the next section, we will generalize the concept of solving \FOND planning problems with extended temporal goals, describing the step by step encoding process of those temporal goals in the \FOND domain, written in \PDDL.
\section{The \FONDFOR approach}\label{sec:fond-plan-appr}
As written in Section \ref{sec:plan-idea-motiv}, planning with extended temporal goals has been considered over the representation of goals in classical planning to capture a richer class of plans where restrictions on the whole sequence of states must be satisfied as well. In particular, differently from classical planning, where the goal description can only be expressed as a propositional formula, in planning for extended temporal goals the goal description may have the same expressive power of the temporal logic in which the goal is specified. This enlarges the general view about planning. In other words, extended temporal goals specify desirable sequences of states and a plan exists if its execution yields one of these desirable sequences \citep{bacchus1998planning}.

In this thesis, we propose a new approach, called \FONDFOR, that uses \LTLf and \PLTLf formalisms as temporal logics for expressing extended goals.
To better understand the powerful of planning with extended temporal goals we give the following example.

\begin{example}\label{ex:pla-temp-simple}
Considering the well known \texttt{triangle tireworld} \FOND planning task. The objective is to drive from one location to another, however while driving a tire may be going flat. If there is a spare tire in the location of the car, then the car can use it to fix the flat tire. The task in depicted in Figure \ref{fig:ttireworld-task}, where there are locations arranged as a triangle, arrows representing roads and circles meaning that in a location there is a spare tire. 

\begin{figure}[h]
\centering
\includegraphics[width=0.5\textwidth]{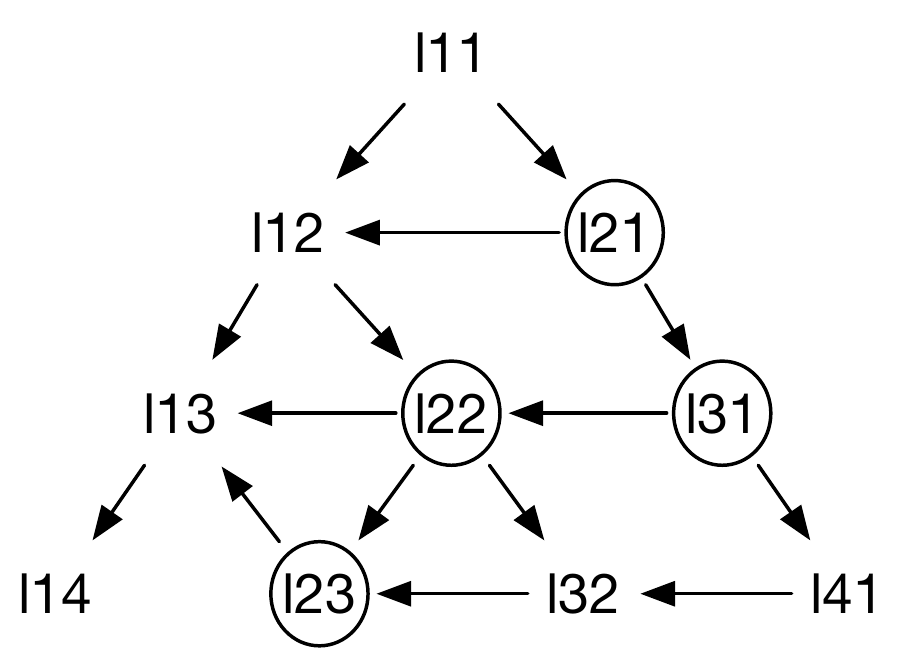}
\caption{A possible Triangle Tireworld task. Locations marked with a circle have a spare tire, arrows represent possible directions} 
\label{fig:ttireworld-task}
\end{figure}

A possible classical goal can be $G = vehicleAt(l31)$, namely a propositional formula saying something only about what have to be true at the end of the execution. In the case of $G$, we exclusively require that the vehicle should be in location $l31$.

On the contrary, a goal specification expressed with temporal formalism such as \PLTLf could be $\varphi = vehicleAt(l13) \lAND \Once(vehicleAt(l23))$. Such a specification requires to reach position $l13$, imposing the passage through position $l23$ before reaching the goal.
\end{example}

Planning for \LTLf and \PLTLf goals slightly changes the definitions given in Section \ref{sec:fond}. In the following, we give the modified definitions of the concepts seen before.

\begin{definition}\label{def:strong-sol-extend}
Given a domain $\D$ and an \LTLf/\PLTLf formula $\varphi$ over atoms $\F \cup \mathcal{A}$, a strategy $\trace$ is a \textit{strong solution} to $\D$ for goal $\varphi$, if every $\trace$-trace is finite and satisfies $\varphi$.
\end{definition}

About complexity of \FONDS, we have the following Theorems.

\begin{theorem}\citep{de2018automata}
\label{th:fond-ltlf-complex}
Solving \FONDS for \LTLf goals is:
\begin{itemize}
\item \EXPTIME-complete in the size of the domain;
\item \TWOEXPTIME-complete in the size of the goal.
\end{itemize}
\end{theorem}

Furthermore, if the goal has the form of $\Diamond G$, i.e. is a reachability goal, the cost with respect to the goal becomes polynomial because it is just a propositional evaluation. If for a given \LTLf goal the determinization step does not cause a state explosion,  the complexity with respect to the goal is \EXPTIME. 
On the contrary, if $G$ is a \PLTLf formula, from results in \cite{de2018automata}, we can only say that the complexity is \TWOEXPTIME in the size of the goal.
Even though we remark that the investigation on planning for \PLTLf goals may have a computational advantage since \PLTLf formulas can be reduced to \DFA in single exponential time (vs. double-exponential time of \LTLf formulas) \citep{chandra1981acm}, computing the hardness is not obvious because we should evaluate the formula only after the $\Diamond$ operator.

\subsection{Idea}
Our \FONDFOR approach works as follows: given a non-deterministic planning domain $\D$, an initial state $s_0$ and an \LTLf or \PLTLf goal formula $\varphi$ (whose symbols are ground predicates), we first obtain the corresponding \DFA of the temporal formula through \LTLfToDFA, then, we encode such a \DFA into the non-deterministic planning domain $\D$. As a result, we will have a new domain $\D'$ and a new problem $P'$ that can be considered and solved as a classical \FOND planning problem. 

The new approach, carried out in this thesis, stems from the research in \cite{de2018automata}, that, basically, proposes automata-theoretic foundations of \FOND planning for \LTLf goals. In particular, they compute the cartesian product between the \DFA corresponding to the domain $\D$ ($\automaton_\D$) and the \DFA corresponding to $\varphi$ ($\automaton_\varphi$), thus, solving a \DFA game on $\automaton_\D \times \automaton_\varphi$, i.e. find, if exists, a trace accepted by $\automaton_\D \times \automaton_\varphi$. 

However, unlike what has been done in \cite{de2018automata}, we split transitions containing the action and its effect in order to have them separately. The reason for this separation is that having both the action and its effect on the same transition is not suitable on a practical perspective. Hence, we have devised a solution in which we run $\automaton_\D$ and $\automaton_\varphi$ separately, but combining them into a single unique transition system. To achieve this, we move $\automaton_\D$ and $\automaton_\varphi$ alternatively by introducing an additional predicate, which we will call \texttt{turnDomain}, that is true when we should move $\automaton_\D$ and is false when we should move $\automaton_\varphi$. In the following, we give an example to better understand the solution put in place in this thesis.

\begin{example}\label{ex:yale-scenario}
Let us consider the simplified version of the classical Yale shooting domain \citep{hanks1986default} as in \cite{de2018automata}, where we have that the turkey is either alive or not and the actions are shoot and wait with the obvious effects, but with a gun that can be faulty. Specifically, shooting with a (supposedly) working gun can either end in killing the turkey or in the turkey staying alive and the discovery that the gun is not working properly. On the other hand, shooting (with care) with a gun that does not work properly makes it work and kills the turkey. The cartesian product $\automaton_\D \times \automaton_\varphi$ with $\varphi = \Diamond\lnot a$ is as follows:

\begin{figure}[h]
\centering
\includegraphics[width=0.7\textwidth]{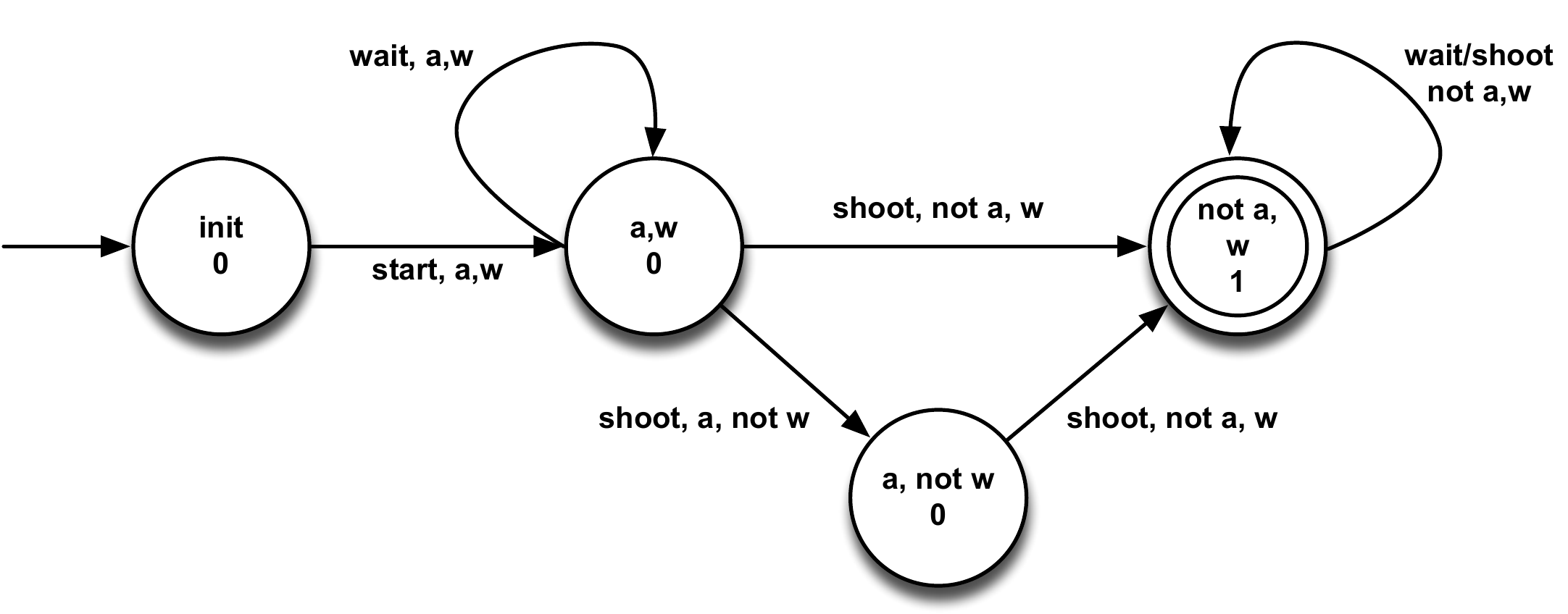}
\caption{The \DFA corresponding to $\automaton_\D \times \automaton_\varphi$. Symbol \texttt{a} stands for \textit{alive} and \texttt{w} for \textit{working}} 
\label{fig:dfa-game}
\end{figure}

\noindent As we can see in Figure \ref{fig:dfa-game}, each transition reads both the action and its effect. This is not suitable for a practical implementation. Thus, we do not perform the cartesian product between the two automata. On the contrast, we build a transition system as in Figure \ref{fig:yale-our-sol}.

\begin{figure}[h]
\centering
\includegraphics[width=0.8\textwidth]{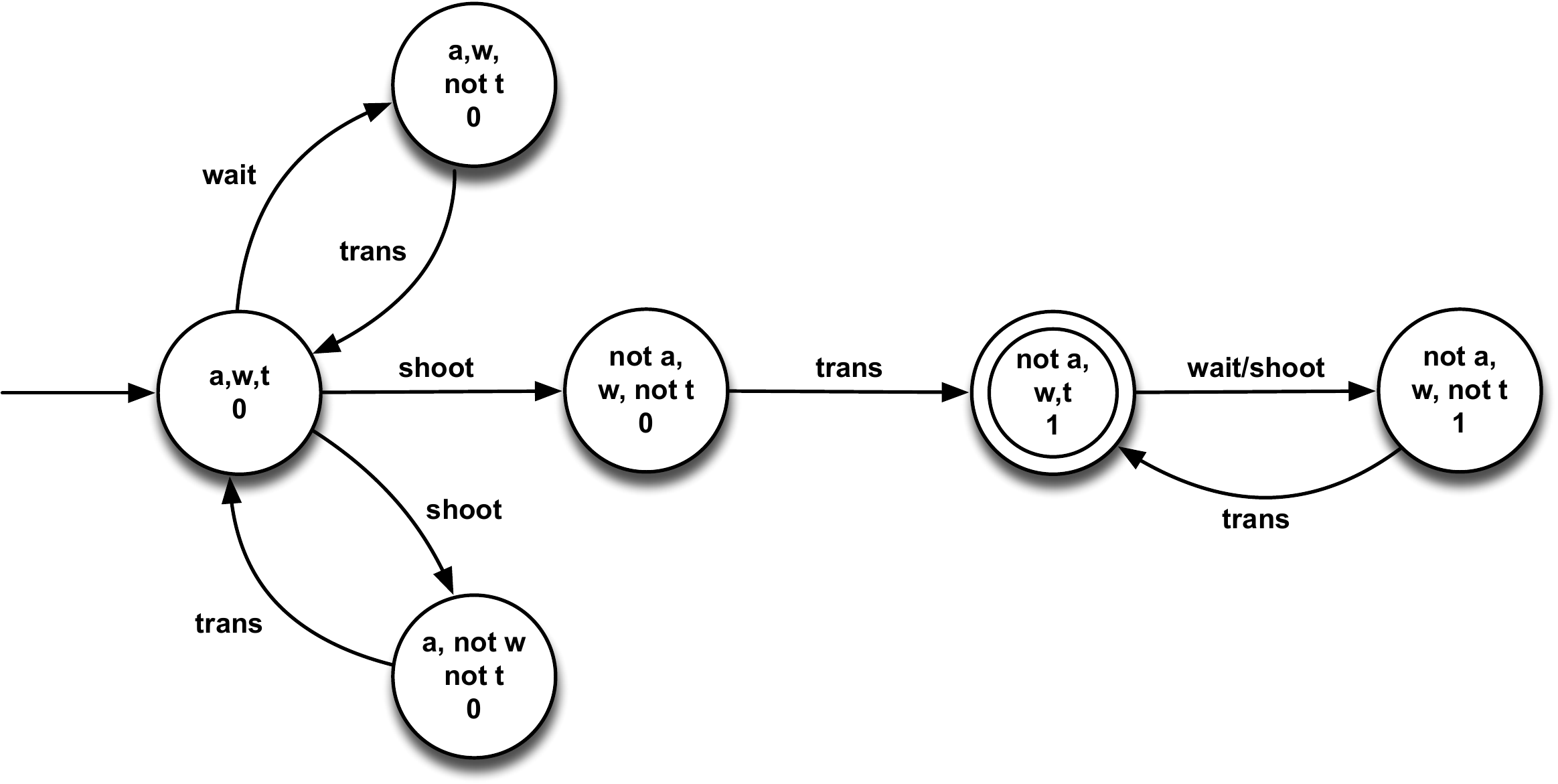}
\caption{The new transition system corresponding to the Yale shooting domain. Symbol \texttt{a} stands for \textit{alive}, \texttt{w} for \textit{working} and \texttt{t} for \textit{turnDomain}} 
\label{fig:yale-our-sol}
\end{figure}
\end{example}

The transition system, shown in Figure \ref{fig:yale-our-sol}, expresses the new domain $\D'$ that has a perfect alternation of transitions. In particular, actions of the initial domain $\D$ alternates with a special action, that we called \texttt{trans}, representing the movement done by $\automaton_\varphi$. Moreover, it is important to notice the usage of the added predicate \texttt{turnDomain} allowing us to alternate between general actions and the new action \texttt{trans}.

In the next section, we will explain how the new domain $\D'$ and the new problem $\P'$ can be written in \PDDL by showing the encoding of \LTLf/\PLTLf goals in \PDDL.

\subsection{Encoding of \LTLf/\PLTLf goals in \PDDL}\label{sec:encoding-goals}
In this section, we describe the process of obtaining the new domain $\D'$ and the new problem $\P'$, both specified in \PDDL. The original \PDDL domain $\D$ and the associated original problem $\P$ change when introducing \LTLf/\PLTLf goals. In particular, what changes is the way we encode our \LTLf/\PLTLf formula in \PDDL. Firstly, we employ our \LTLfToDFA tool to convert the given goal formula $\varphi$ into the corresponding \DFA. Then, we encode, in a specific way, the resulting \DFA automaton in \PDDL modifying the original domain $\D$ and problem $\P$.

\subsubsection*{Translation of \DFAs in \PDDL in Domain $\D$}
In order to explain the translation technique of a \DFA to \PDDL, we assume to already have the \DFA generated by \LTLfToDFA. We recall that such a \DFA is formally defined as follows:
\begin{definition}\label{plan:dfa}
A \DFA is a tuple $\automaton = \tup{\Sigma, Q, q_0 , \delta, F}$, where:
\begin{itemize}		
\item $\Sigma = \{ a_0,a_1,\dots,a_n \}$ is a finite set of symbols;
\item $Q = \{ q_0,q_1,\dots,q_m \}$ is the finite set of states;
\item $q_0 \in Q$ is the initial state;
\item $\delta: Q \times \Sigma \rightarrow Q$ is the transition function;
\item $F \subseteq Q$ is the set of final states;
\end{itemize}
\end{definition}
Specifically, since the automaton $\automaton$ corresponds to the goal formula $\varphi$, which has grounded predicates as symbols, we can represent them as $\{a_0(o_0,\dots,o_j),\dots, a_n(o_0,\dots,o_w)\}$, where $o_0,\dots,o_k \in \O$ and $0 \leq j,w \leq k$ represents objects 
present in the problem $\P$.
To capture the general representation of $\varphi$ in $\D$, we have to modify $\automaton$ to $\hat{\automaton}$ performing a transformation explained below. We give the following definitions.

\begin{definition}\label{plan:automaton-prime}
$\hat{\automaton}$ is a tuple $\hat{\automaton} = \tup{\hat{\Sigma}, \hat{Q}, \hat{q}_0 , \hat{\delta}, \hat{F}}$, where:
\begin{itemize}		
\item $\hat{\Sigma} = \{ \hat{a}_0,\hat{a}_1,\dots,\hat{a}_n \}$ is a finite set of symbols;
\item $\hat{Q} = \{ \hat{q}_0,\hat{q}_1,\dots,\hat{q}_m \}$ is the finite set of states;
\item $\hat{q}_0 \in \hat{Q}$ is the initial state;
\item $\hat{\delta}: \hat{Q} \times \hat{\Sigma} \rightarrow \hat{Q}$ is the transition function;
\item $\hat{F} \subseteq \hat{Q}$ is the set of final states;
\end{itemize}
\end{definition}

\begin{definition}\label{plan:mapfunc}
Given the set of \DFA symbols $\Sigma$, we define a mapping function $m$ as follows:
\begin{equation}
m: \O \rightarrow \V
\end{equation}
where $\O$ is the set of objects $\{o_0,\dots,o_k\}$ and $\V$ is a set of variables $\{x_0,\dots,x_k\}$
\end{definition}

The transformation from $\automaton$ and $\hat{\automaton}$ is carried out with the mapping function $m$ as follows:
\begin{itemize}
\item $\hat{\Sigma} = \{ \hat{a}_0,\dots,\hat{a}_n \}$, where $\hat{a}_i \doteq a_i(x_0,\dots,x_j)$ and $x_0,\dots,x_j \subseteq \V$;
\item $\hat{Q} = \{ \hat{q}_0,\dots,\hat{q}_m \}$, where $\hat{q}_i \doteq q_i(x_0\dots, x_k)$.
\end{itemize}
Then, $\hat{q}_0, \hat{\delta} \text{ and } \hat{F}$ are modified accordingly.

Once the transformation is done, we have obtained a \emph{parametric} \DFA, which is a general representation with respect to the original one. After that, for representing the \DFA transitions in the domain $\D$, we should encode the new \textit{transition function} $\hat{\delta}$ into \PDDL. To this extent, the $\hat{\delta}$ function is represented as a new \PDDL operator, called \texttt{trans} having  these properties:

\begin{itemize}
\item all variables in $\V$ are parameters;
\item the negation of the \texttt{turnDomain} predicate is a precondition;
\item effects represent the $\hat{\delta}$ function.
\end{itemize}
Moreover, effects are expressed as conditional effects. The general encoding would be as follows:
\begin{lstlisting}[escapechar=£]
Action £\texttt{trans}£:
  parameters: £$(x_0, \dots, x_k)$, where $x_i \in \V$£
  preconditions: £$\lnot turnDomain$£
  effects: £when $(q_i(x_0, \dots, x_k) \lAND \hat{a}_j)$ then $(\hat{\delta}(\hat{q}_i, \hat{a}_j)=q'_i(x_0, \dots, x_k) \lAND (\lnot q, \forall q \in \hat{Q} \text{ s.t. } q \neq q'_i) \lAND turnDomain),\; \forall i,j: 0\leq i \leq m, 0\leq j \leq n$£
\end{lstlisting}

Additionally, in \PDDL, especially in the effect formula of a conditional effect, we should specify that if the automaton is in a state, it is not in other states. This is captured by adding the negation of all other automaton states. In the following, we give an example showing the translation of \DFAs to \PDDL step-by-step.

\begin{example}\label{ex:param-formula}
Let us consider the goal formula $\varphi = \Diamond(on(d3,rod3))$ for the Tower of Hanoi planning problem. The predicate \texttt{on} is instantiated on objects \texttt{d3} and \texttt{rod3}. By applying the mapping function $m$ we have the corresponding variables and $\varphi$ becomes $\varphi(x_1,x_2) = \Diamond(on(x_1,x_2))$, where we know that $x_1 = f(d_3)$ and  $x_2 = f(rod_3)$. In this case, the modified \DFA $\automaton'$ is depicted in Figure \ref{fig:dfa-parametric}.
\begin{figure}[h]
\centering
\includegraphics[width=0.7\textwidth]{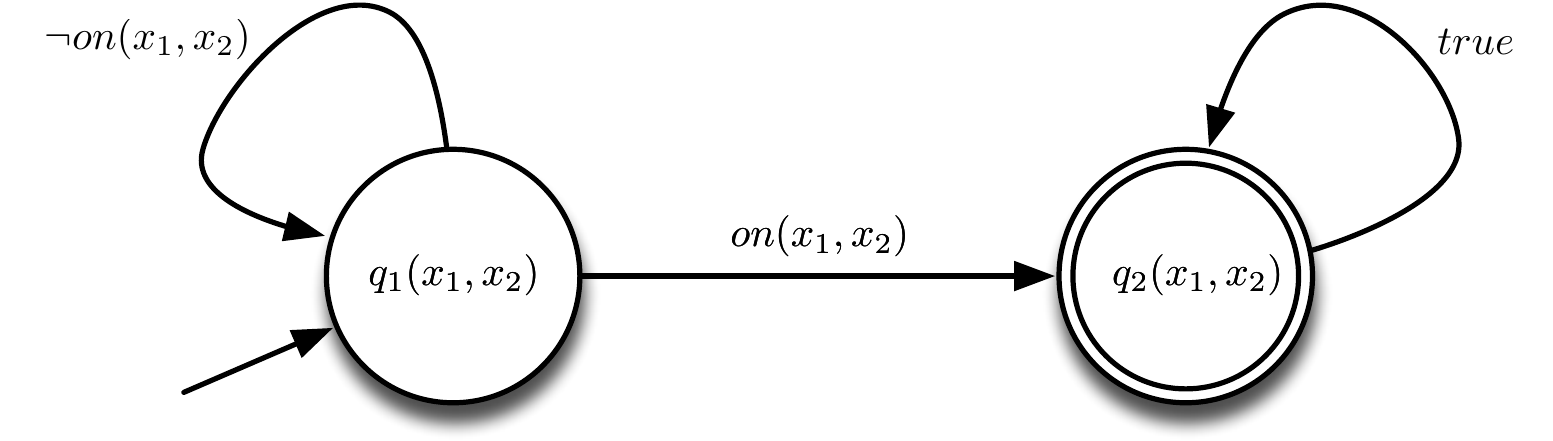}
\caption{The parametric \DFA corresponding to $\varphi(x_1,x_2) = \Diamond(on(x_1,x_2))$} 
\label{fig:dfa-parametric}
\end{figure}
At this point, consider the new \DFA, the \texttt{trans} operator built from that automaton is the following:
\begin{lstlisting}[language=PDDL, escapechar=£]
(:action trans
  :parameters (?x1 ?x2)
  :precondition (not (turnDomain))
  :effect (and (when (and (q1 ?x1 ?x2) (not (on ?x1 ?x2)))
              (and (q1 ?x1 ?x2) (not (q2 ?x1 ?x2)) (turnDomain))
          (when (or (and (q1 ?x1 ?x2) (on ?x1 ?x2)) (q2 ?x1 ?x2))£\label{line:cond-eff-ex}£
              (and (q2 ?x1 ?x2) (not (q1 ?x1 ?x2)) (turnDomain))
  )
)
\end{lstlisting}
\end{example}

As just shown in Example \ref{ex:param-formula}, transitions, with source state and destination state, are encoded as conditional effects, where the condition formula includes source state and formula symbols whereas the effect formula includes the destination state, the negation of all other states and \texttt{turnDomain}. Moreover, in order to get a compact encoding of \texttt{trans} effects, conditional effects are brought together by destination state as happens, for instance, at line \ref{line:cond-eff-ex}.

After the \texttt{trans} operator has been built, we change $\D$ as follows:
\begin{enumerate}
\item $\forall a \in A$: add \texttt{turnDomain} to $\alpha(s)$, i.e add \texttt{turnDomain} predicate to all actions precondition $\alpha(s)$;
\item $\forall a \in A$: add \texttt{(not (turnDomain))} to $(s,a,s') \in \varrho$ with $a \in \alpha(s)$, i.e add negated \texttt{turnDomain} to all actions effects $(s,a,s')$;
\item add \texttt{trans} operator;
\item $\forall q' \in Q'$: add $q'$ to predicates definition of $\D$, i.e. add all automaton state predicates to the domain predicates definition.
\end{enumerate}

\noindent We have thus obtained the new domain $\D'$. In the following, we show an example.

\begin{example}\label{ex:new-dom}
Let consider again the Triangle Tireworld scenario. The original \PDDL domain is:
\begin{lstlisting}[language=PDDL, escapechar=£]
(define (domain triangle-tire)
  (:requirements :typing :strips :non-deterministic)
  (:types location)
  (:predicates (vehicle-at ?loc - location)
	       (spare-in ?loc - location)
	       (road ?from - location ?to - location)
	       (not-flattire))
  (:action move-car
    :parameters (?from - location ?to - location)
    :precondition (and (vehicle-at ?from) (road ?from ?to) 
      (not-flattire))
    :effect (oneof (and (vehicle-at ?to) (not (vehicle-at ?from)))
	  (and (vehicle-at ?to) (not (vehicle-at ?from)) 
	  (not (not-flattire))))
   )
  (:action changetire
    :parameters (?loc - location)
    :precondition (and (spare-in ?loc) (vehicle-at ?loc))
    :effect (and (not (spare-in ?loc)) (not-flattire))
   )
)
\end{lstlisting}
Now, consider a simple \LTLf formula $\varphi = \Diamond vehicleAt(l13)$. It requires that eventually the vehicle will be in location $l13$. The parametric \DFA associated to $\varphi(x)$ is depicted in Figure \ref{fig:dfa-parametric2}.
\begin{figure}[h]
\centering
\includegraphics[width=0.7\textwidth]{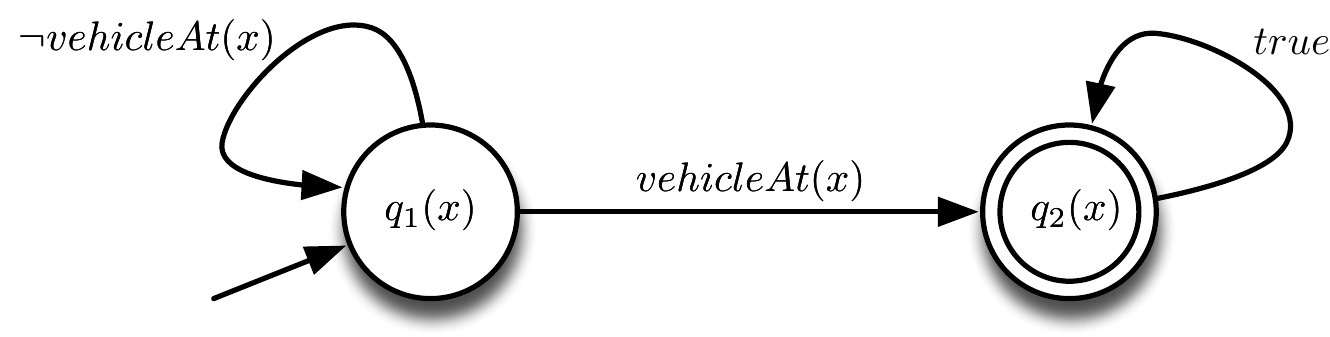}
\caption{The parametric \DFA corresponding to $\varphi(x) = \Diamond(vehicleAt(x))$} 
\label{fig:dfa-parametric2}
\end{figure}

Considering the \DFA in Figure \ref{fig:dfa-parametric2}, the \texttt{trans} operator built from that automaton is the following:
\begin{lstlisting}[language=PDDL, escapechar=£]
(:action trans
  :parameters (?x - location)
  :precondition (not (turnDomain))
  :effect (and (when (and (q1 ?x) (not (vehicle-at ?x)))
              (and (q1 ?x) (not (q2 ?x)) (turnDomain))
          (when (or (and (q1 ?x) (vehicle-at ?x)) (q2 ?x))
              (and (q2 ?x) (not (q1 ?x)) (turnDomain))
  )
)
\end{lstlisting}

Finally, putting together all pieces and carrying out changes described above, we obtain the new domain $\D'$ as follows:

\begin{lstlisting}[language=PDDL, escapechar=£]
(define (domain triangle-tire)
  (:requirements :typing :strips :non-deterministic)
  (:types location)
  (:predicates (vehicle-at ?loc - location)
	       (spare-in ?loc - location)
	       (road ?from - location ?to - location)
	       (not-flattire)
	       (q1 ?x - location)
	       (q2 ?x - location)
	       (turnDomain))
  (:action move-car
    :parameters (?from - location ?to - location)
    :precondition (and (vehicle-at ?from) (road ?from ?to) 
      (not-flattire) (turnDomain))
    :effect (oneof (and (vehicle-at ?to) (not (vehicle-at ?from))
      (not (turnDomain)))
	  (and (vehicle-at ?to) (not (vehicle-at ?from)) 
	  (not (not-flattire)) (not (turnDomain))))
   )
  (:action changetire
    :parameters (?loc - location)
    :precondition (and (spare-in ?loc) (vehicle-at ?loc) 
      (turnDomain))
    :effect (and (not (spare-in ?loc)) (not-flattire)
      (not (turnDomain)))
   )
  (:action trans
    :parameters (?x - location)
    :precondition (not (turnDomain))
    :effect (and (when (and (q1 ?x) (not (vehicle-at ?x)))
              (and (q1 ?x) (not (q2 ?x)) (turnDomain))
          (when (or (and (q1 ?x) (vehicle-at ?x)) (q2 ?x))
              (and (q2 ?x) (not (q1 ?x)) (turnDomain)))
  )   
)
\end{lstlisting}
\end{example}

\subsubsection*{Change in Problem $\P$}
Concerning the planning problem $\P$, we completely discard the goal specification, whereas the initial state description is slightly modified. Moreover, the problem name, the associated domain name and all defined objects remain unchanged. We have to modify both the initial state and the goal state specifications to make them compliant with $\D'$, containing all changes introduced in the planning domain $\D$. To this extent, we formally define the new initial state as follows:
\begin{equation}
\text{Init: } s_0 \lAND \text{\texttt{turnDomain}} \lAND q_0
\end{equation}
where $q_0 \doteq \hat{q}_0(m^{-1}(x_0),\dots,m^{-1}(x_k)) = q_0(o_0,\dots,o_k)$. In other words, we put together the original initial specification $s_0$, the new predicate \texttt{turnDomain}, meaning that it is $true$ at the beginning, and the initial state of the automaton instantiated on the objects of interest, i.e. those specified in the \LTLf/\PLTLf formula $\varphi$. 

On the other hand, the goal description is built from scratch as follows:
\begin{equation}
\text{Goal: } \text{\texttt{turnDomain}} \lAND (\bigvee_{q \in F} q)
\end{equation}
where $q \doteq \hat{q}(m^{-1}(x_0),\dots,m^{-1}(x_k)) = q(o_0,\dots,o_k)$. In other words, we place  together the \texttt{turnDomain} predicate, meaning that it must be $true$ at the end of the execution, and the final state(s) of the automaton always instantiated on the objects of interest. Here, it is important to notice that if the automaton has two or more final states, they should be put in disjunction.

We can give the following example.

\begin{example}\label{ex:new-prob}
Let consider again the Triangle Tireworld scenario, shown in the Example \ref{ex:new-dom}. The original \PDDL domain is:
\begin{lstlisting}[language=PDDL, escapechar=£]
(define (problem triangle-tire-1)
  (:domain triangle-tire)
  (:objects l11 l12 l13 l21 l22 l23 l31 l32 l33 - location)
  (:init (vehicle-at l11)
    (road l11 l12) (road l12 l13) (road l11 l21) (road l12 l22)
    (road l21 l12) (road l22 l13) (road l21 l31) (road l31 l22)
    (spare-in l21) (spare-in l22) (spare-in l31)
    (not-flattire))
  (:goal (vehicle-at l13))
)
\end{lstlisting}
Now, considering the same \LTLf formula $\varphi = \Diamond vehicleAt(l13)$, the object of interest is $l13$. Hence, we should evaluate our automaton states as $q_1(l13)$ and $q_2(l13)$.

Finally, putting together all pieces and carrying out changes described above, we obtain the new problem $\P'$ as follows:

\begin{lstlisting}[language=PDDL, escapechar=£]
(define (problem triangle-tire-1)
  (:domain triangle-tire)
  (:objects l11 l12 l13 l21 l22 l23 l31 l32 l33 - location)
  (:init (vehicle-at l11)
    (road l11 l12) (road l12 l13) (road l11 l21) (road l12 l22)
    (road l21 l12) (road l22 l13) (road l21 l31) (road l31 l22)
    (spare-in l21) (spare-in l22) (spare-in l31)
    (not-flattire) (turnDomain) (q1 l13))
  (:goal (and (turnDomain) (q2 l13)))
)
\end{lstlisting}
As a remark, we will refer to the new goal specification as $\G'$.
\end{example}

Having examined the encoding of \LTLf/\PLTLf goal formulas in \PDDL, the resulting planning domain $\D'$ and problem $\P'$ represent a ``classical'' planning specification. In the next Section, we will see how we obtain a strong policy giving $\D'$ and $\P'$.

\subsection{\FOND Planners}
In this Section, we talk about the state-of-art \FOND planners and how they are employed within this thesis.

To begin with, thanks to our encoding process, we have reduced the problem of \FOND planning for \LTLf/\PLTLf goals to a ``classical'' \FOND planning, which is essentially a \textit{reachability} problem. We can state the following Theorem.

\begin{theorem}
A strong policy $\trace$ is a valid policy for $\D', \G'$ if and only if $\trace$ is a valid policy for $\D,\varphi$.
\end{theorem}

Given this Theorem, we can solve our original problem giving $\D'$ and $\P'$ as input to standard \FOND planners. The main state-of-art \FOND planners are:
\begin{itemize}
\item MBP and Gamer, which are OBDD\footnote{OBDD stands for \textit{Ordered Binary Decision Diagram}}-based planners \citep{cimatti2003weak, kissmann2009solving}
\item MyND and Grendel, which rely on explicit AND/OR graph search \citep{bercher2010pattern, ramirez2014directed}
\item PRP, NDP and FIP, which rely on classical algorithms \citep{kuter2008using, fu2011simple, muise2012improved}
\item FOND-SAT, which provides a SAT approach to \FOND planning \citep{geffner2018compact}
\end{itemize}
Although \FOND planning is receiving an increasingly interest, the research on computational approaches has been recently reduced. Nevertheless, some planners performs well on different contexts of use. In our thesis, we are going to employ a customized version of FOND-SAT, the newest planner.

Secondly, although the description of many real-world planning problems involves the use of conditional effects, requiring full support of ADL by planners, those state-of-art planners, cited above, are still not able to fully handle such conditional effects. This represents a big limitation that can be surely deepened as a future work of this thesis. To this extent, we should first compile away conditional effects from the domain and, then, we can give it to a planner. Our proposal implementation is able to compile away simple conditional effects, namely those conditional effects that do not have nested formulas. Additionally, we have to  compile away conditional effects of the \texttt{trans} operator upstream even though its  representation with the employment of conditional effects is much more effective and compact. Luckily, this process consists of just splitting the operator in as many operators as the number of conditional effects present in the original action and adding the condition formula in the preconditions for each conditional effect. In the following, we make a clarifying example.

\begin{example}
The \texttt{trans} operator built in the Example \ref{ex:new-dom} is:
\begin{lstlisting}[language=PDDL, escapechar=£]
(:action trans
  :parameters (?x - location)
  :precondition (not (turnDomain))
  :effect (and (when (and (q1 ?x) (not (vehicle-at ?x)))
              (and (q1 ?x) (not (q2 ?x)) (turnDomain))
          (when (or (and (q1 ?x) (vehicle-at ?x)) (q2 ?x))
              (and (q2 ?x) (not (q1 ?x)) (turnDomain))
  )
)
\end{lstlisting}
As we can see, it contains only two conditional effects. Hence, we split this operator in two operators that we are going to call \texttt{trans-0} and \texttt{trans-1}, respectively. In particular, for each conditional effect the condition formula is added to the precondition and the effect formula is left in the effects. \texttt{trans-0} and \texttt{trans-1} are as follows:
\begin{lstlisting}[language=PDDL, escapechar=£]
(:action trans-0
  :parameters (?x - location)
  :precondition (and (and (q1 ?x) (not (vehicle-at ?x))) 
    (not (turnDomain)))
  :effect (and (q1 ?x) (not (q2 ?x)) (turnDomain))
  )
)
(:action trans-1
  :parameters (?x - location)
  :precondition (and (or (and (q1 ?x) (vehicle-at ?x)) (q2 ?x)) 
    (not (turnDomain)))
  :effect (and (q2 ?x) (not (q1 ?x)) (turnDomain))
  )
)
\end{lstlisting}
\end{example}

At this point, having compiled away simple conditional effects from the modified domain $\D'$, we can finally describe how we have employed FOND-SAT in our thesis.

The main reasons why we have chosen FOND-SAT are the following:
\begin{itemize}
\item it is written in pure Python, hence we can easily integrate it in our Python implementation;
\item it performs reasonably well;
\item it outputs all policies building a transition system whose states are called \textit{controller states}.
\end{itemize}
FOND-SAT takes also advantage of the parser and translation PDDL-to-SAS+ scripts from  PRP. When FOND-SAT was developed, PRP's translation scripts could not handle disjunctive preconditions that may be present in our \texttt{trans} operators. As a result, we have modified FOND-SAT with the newest version of those scripts directly from PRP.

The usage of FOND-SAT is really simple. From its source folder, it is necessary to run the following command in the terminal:
\begin{lstlisting}[language=bash]
python main.py -strong 1 -policy 1 /path-to/domain.pddl 
/path-to/problem.pddl
\end{lstlisting}
The command simply executes the main module of FOND-SAT requiring to find strong policies and to print, if exists, the policy found. We feed FOND-SAT with our new domain $\D'$ and new problem $\P'$.

Once strong plans are found, FOND-SAT displays the policy in four sections as follows:
\begin{itemize}
\item Atom (CS): for each controller state it tells what predicates are true;
\item (CS, Action with arguments): for each controller state it tells which actions can be applied
\item (CS, Action name, CS): it tells for each controller state what action is applied in that state and the successor state;
\item (CS1, CS2): it means that the controller can go from CS1 to CS2. 
\end{itemize}
Now, we give an output example.

\begin{example}
The following result has been obtained running FOND-SAT with the Triangle Tireworld domain and problem with the \LTLf goal $\varphi = \Diamond vehicleAt(l31)$. What follows is only the displayed policy.
\begin{lstlisting}[numberstyle=\tiny\color{codegray}\noncopynumber,numbers=left,stepnumber=1, escapechar=£]
...
Trying with 7 states
Looking for strong plans: True
Fair actions: True
# Atoms: 18
# Actions: 26
SAT formula generation time = 0.052484
# Clauses = 11041
# Variables = 1225
Creating formula...
Done creating formula. Calling solver...
SAT solver called with 4096 MB and 3599 seconds
Done solver. Round time: 0.016456
Cumulated solver time: 0.055322
===================
===================
Controller -- CS = Controller State
===================
===================
Atom (CS)
___________________
----------
Atom q1(l31) (n0)
Atom vehicleat(l11) (n0)
Atom not-flattire() (n0)
Atom spare-in(l21) (n0)
Atom turndomain() (n0)
----------
-NegatedAtom turndomain() (n1)
Atom q1(l31) (n1)
Atom vehicleat(l21) (n1)
Atom not-flattire() (n1)
----------
Atom q1(l31) (n2)
-NegatedAtom turndomain() (n2)
Atom spare-in(l21) (n2)
Atom vehicleat(l21) (n2)
----------
Atom turndomain() (n3)
Atom q1(l31) (n3)
Atom vehicleat(l21) (n3)
Atom not-flattire() (n3)
----------
Atom q1(l31) (n4)
Atom spare-in(l21) (n4)
Atom vehicleat(l21) (n4)
Atom turndomain() (n4)
----------
Atom q1(l31) (n5)
Atom vehicleat(l31) (n5)
-NegatedAtom turndomain() (n5)
----------
Atom turndomain() (ng)
Atom q2(l31) (ng)
===================
===================
(CS, Action with arguments)£\label{line:cs-actarg}£
___________________
(n0,move-car_DETDUP_0(l11,l21))
(n0,move-car_DETDUP_1)
(n0,move-car_DETDUP_1(l11,l21))
(n0,move-car_DETDUP_0)
(n1,trans-0_v4)
(n1,trans-0_v4(l31))
(n2,trans-0_v4(l31))
(n2,trans-0_v4)
(n3,move-car_DETDUP_0(l21,l31))
(n3,move-car_DETDUP_0)
(n3,move-car_DETDUP_1(l21,l31))
(n3,move-car_DETDUP_1)
(n4,changetire(l21))
(n4,changetire)
(n5,trans-11)
(n5,trans-11(l31))
===================
===================
(CS, Action name, CS)
___________________
(n0,move-car_DETDUP_0,n1)£\label{line:cs-actarg}£
(n0,move-car_DETDUP_1,n2)
(n1,trans-0_v4,n3)
(n2,trans-0_v4,n4)
(n3,move-car_DETDUP_0,n5)
(n3,move-car_DETDUP_1,n5)
(n4,changetire,n1)
(n5,trans-11,ng)
===================
(CS, CS)
___________________
(n2,n4)
(n5,ng)
(n3,n5)
(n1,n3)
(n0,n2)
(n4,n1)
(n0,n1)
===================
Solved with 7 states
Elapsed total time (s): 0.288035
Elapsed solver time (s): 0.055322
Elapsed solver time (s): [0.0052, 0.0059, 0.007, 0.009, 0.011, 0.016]
Looking for strong plans: True
Fair actions: True
Done

\end{lstlisting}
As we can see, operators names are changed due to internal arrangements made by FOND-SAT needed to handle both non determinism and disjunctive preconditions. Here, it is important to observe that, as we expected, there is an alternation of action executions between the original domain actions and the \texttt{trans} operators. Finally, the transition system built by FOND-SAT has $n0$ as initial state and $ng$ as final state. If strong plans are found, it means that every path from $n0$ to $ng$ is a valid plan. We will better explain this later in Section \ref{sec:planning-results}.
\end{example}

In the following section, we will report results obtained through our practical implementation, called \FONDFOR, that automates all the process illustrated in this Section.

\section{Results}\label{sec:planning-results}
In this section, we show an execution of the \FONDFOR tool as example of result. In particular, we show an execution involving a \PLTLf goal. We go step-by-step through the solution and we pay  attention to the \texttt{trans} operator.

\begin{figure}[h]
\centering
\includegraphics[width=0.58\textwidth]{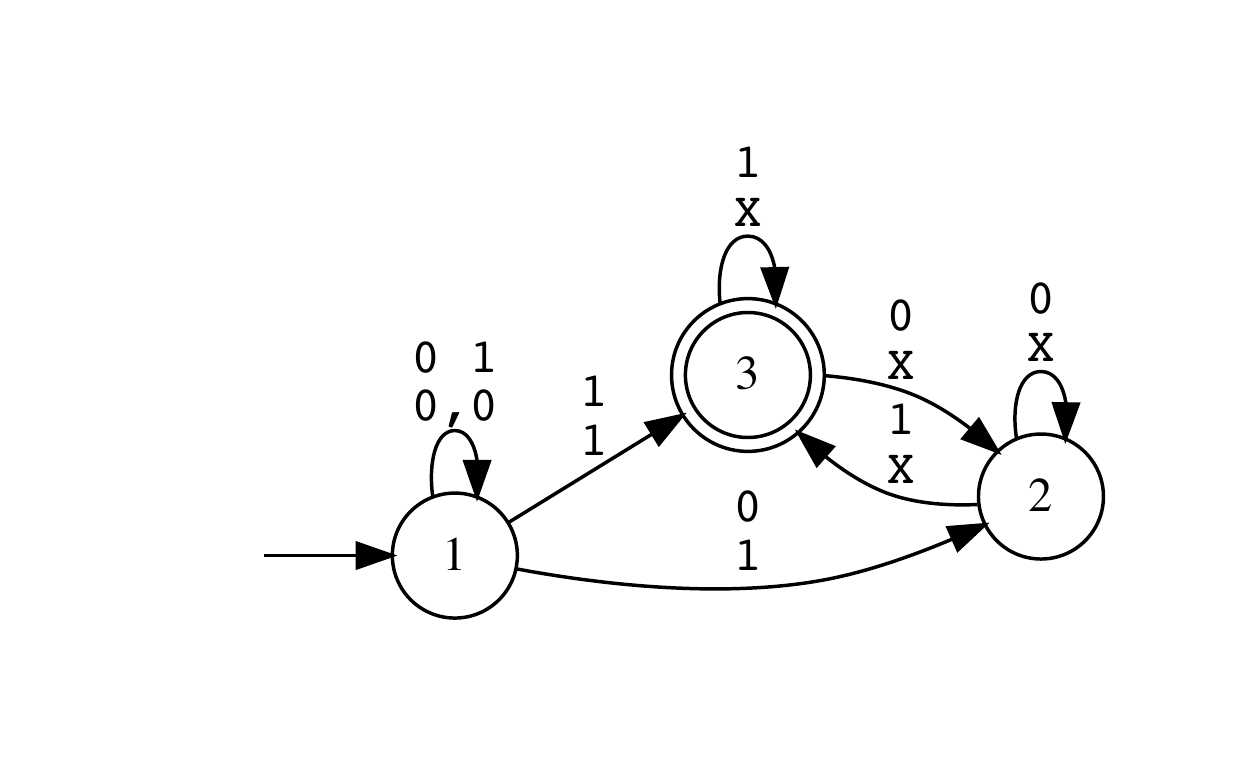}
\caption{The \DFA corresponding to $\varphi$} 
\label{fig:dfa-result-pltl}
\end{figure}

We remind the reader that \FONDS planning for \PLTLf goals is interpreted as reaching a final state such that the history leading to such a state satisfies the given \PLTLf formula.
For instance, here we show an execution of the \FONDFOR tool on the Triangle Tireworld planning task partly illustrated in Example \ref{ex:new-dom}. Indeed, the original domain is the same of the one in Example \ref{ex:new-dom} whereas the original initial state is as follows:
\begin{lstlisting}[language=PDDL, escapechar=£]
(define (problem triangle-tire-1)
  (:domain triangle-tire)
  (:objects l11 l12 l13 l21 l22 l23 l31 l32 l33 - location)
  (:init (vehicle-at l11)
    (road l11 l12) (road l12 l13) (road l11 l21) (road l12 l22)
    (road l21 l12) (road l22 l13) (road l21 l31) (road l31 l22)
    (spare-in l21) (spare-in l22) (spare-in l31)
    (not-flattire))
  (:goal (vehicle-at l13))
)
\end{lstlisting}
In this case, we choose the \PLTLf goal formula $\varphi = vehicleAt(l13) \lAND \Once(vehicleAt(l23))$. Such a formula means \emph{reach location l13 passing through location l23 at least once}. The \DFA corresponding to $\varphi$, generated with \LTLfToDFA, is depicted in Figure \ref{fig:dfa-result-pltl}.

After the execution of \FONDFOR, we obtain the following:
\begin{enumerate}
\item a new planning domain;
\item a new planning problem;
\item a transition system showing all policies, if found.
\end{enumerate}
Firstly, the new domain $\D'$ is:
\begin{lstlisting}[language=PDDL, escapechar=£]
(define (domain triangle-tire)
  (:requirements :typing :strips :non-deterministic)
  (:types location)
  (:predicates (vehicleat ?loc - location) (spare-in ?loc - location) 
  (road ?from - location ?to - location) (not-flattire) (turnDomain) 
  (q2 ?loc30 - location ?loc53 - location) 
  (q1 ?loc30 - location ?loc53 - location) 
  (q3 ?loc30 - location ?loc53 - location))
  (:action move-car
    :parameters (?from - location ?to - location)
    :precondition (and (vehicleat ?from) (road ?from ?to) 
    (not-flattire) (turnDomain))
    :effect (and (oneof (and (vehicleat ?to) 
    (not (vehicleat ?from))) (and (vehicleat ?to) 
    (not (vehicleat ?from)) (not (not-flattire)))) 
    (not (turnDomain)))
  )
  (:action changetire
    :parameters (?loc - location)
    :precondition (and (spare-in ?loc) (vehicleat ?loc) 
    (turnDomain))
    :effect (and (not (spare-in ?loc)) (not-flattire) 
    (not (turnDomain)))
  )
  (:action trans-0
    :parameters (?loc30 - location ?loc53 - location)
    :precondition (and (or (and (q1 ?loc30 ?loc53) 
    (not (vehicleat ?loc30)) (vehicleat ?loc53)) 
    (and (q2 ?loc30 ?loc53) (not (vehicleat ?loc30))) 
    (and (q3 ?loc30 ?loc53) (not (vehicleat ?loc30)))) 
    (not (turnDomain)))
    :effect (and (q2 ?loc30 ?loc53) 
    (not (q1 ?loc30 ?loc53)) (not (q3 ?loc30 ?loc53)) 
    (turnDomain))
  )
  (:action trans-1
    :parameters (?loc30 - location ?loc53 - location)
    :precondition (and (or (and (q1 ?loc30 ?loc53) 
    (not (vehicleat ?loc30)) (not (vehicleat ?loc53))) 
    (and (q1 ?loc30 ?loc53) (vehicleat ?loc30) 
    (not (vehicleat ?loc53)))) (not (turnDomain)))
    :effect (and (q1 ?loc30 ?loc53) 
    (not (q2 ?loc30 ?loc53)) (not (q3 ?loc30 ?loc53)) 
    (turnDomain))
  )
  (:action trans-2
    :parameters (?loc30 - location ?loc53 - location)
    :precondition (and (or (and (q1 ?loc30 ?loc53) 
    (vehicleat ?loc30) (vehicleat ?loc53)) 
    (and (q2 ?loc30 ?loc53) (vehicleat ?loc30)) 
    (and (q3 ?loc30 ?loc53) (vehicleat ?loc30))) 
    (not (turnDomain)))
    :effect (and (q3 ?loc30 ?loc53) 
    (not (q2 ?loc30 ?loc53)) (not (q1 ?loc30 ?loc53)) 
    (turnDomain))
  )
)
\end{lstlisting}

Secondly, the new problem $\P'$ is:
\begin{lstlisting}[language=PDDL, escapechar=£]
(define (problem triangle-tire-1)
	(:domain triangle-tire)
	(:objects l11 l12 l13 l21 l22 l23 l31 l32 l33 - location)
	(:init (not-flattire) (q1 l13 l23) (road l11 l12)(road l11 l21) 
	(road l12 l13) (road l12 l22) (road l21 l12) (road l21 l31) 
	(road l22 l13) (road l22 l23) (road l23 l13) (road l31 l22) 
	(spare-in l21) (spare-in l22) (spare-in l23) (spare-in l31)
	(turnDomain) (vehicleat l11))
(:goal (and (q3 l13 l23) (turnDomain)))
)
\end{lstlisting}

\begin{figure}[h]
\centering
\includegraphics[scale=.8]{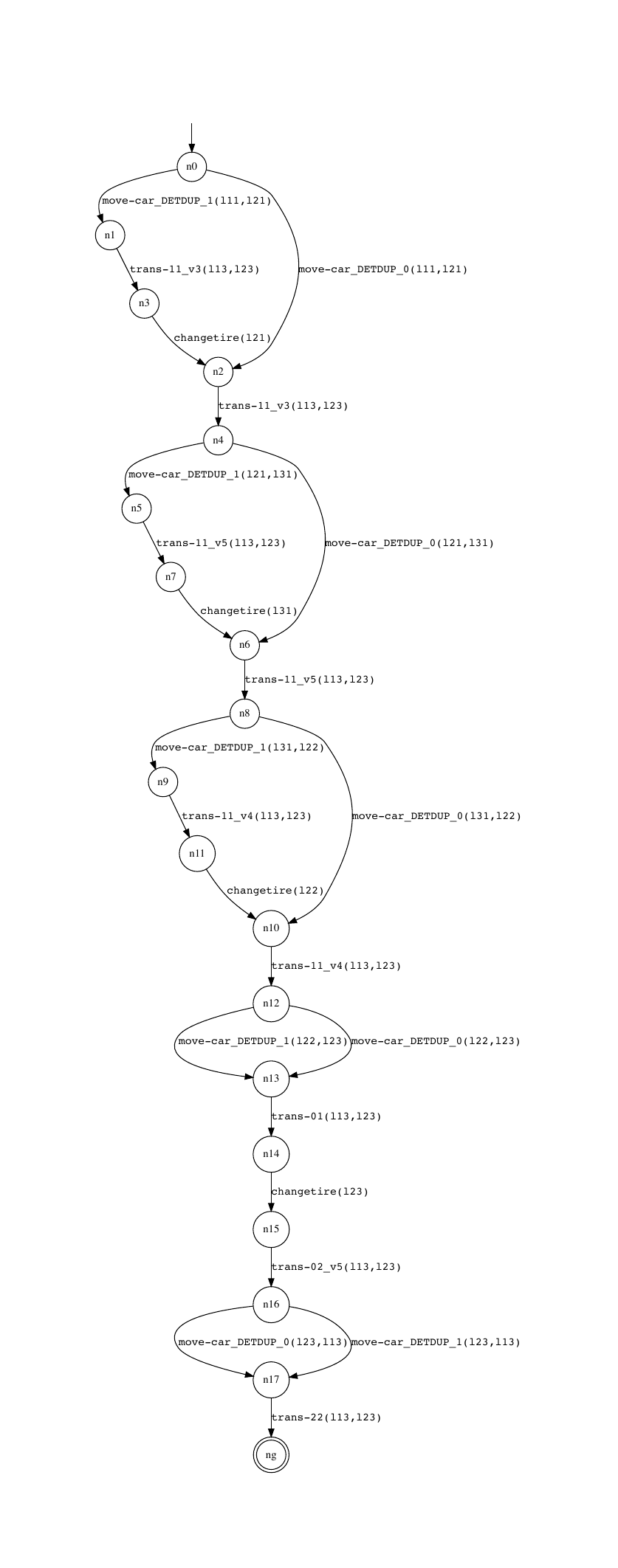}
\caption{The transition system showing policies found with the \texttt{trans} operator} 
\label{fig:policy-trans}
\end{figure}

Finally, feeding FOND-SAT with $\D'$ and $\P'$ we obtain the following transition system depicted in Figure \ref{fig:policy-trans}, thanks to another Python script, developed in this thesis, for converting a written policy into a graph.

\begin{figure}[h]
\centering
\includegraphics[scale=.8]{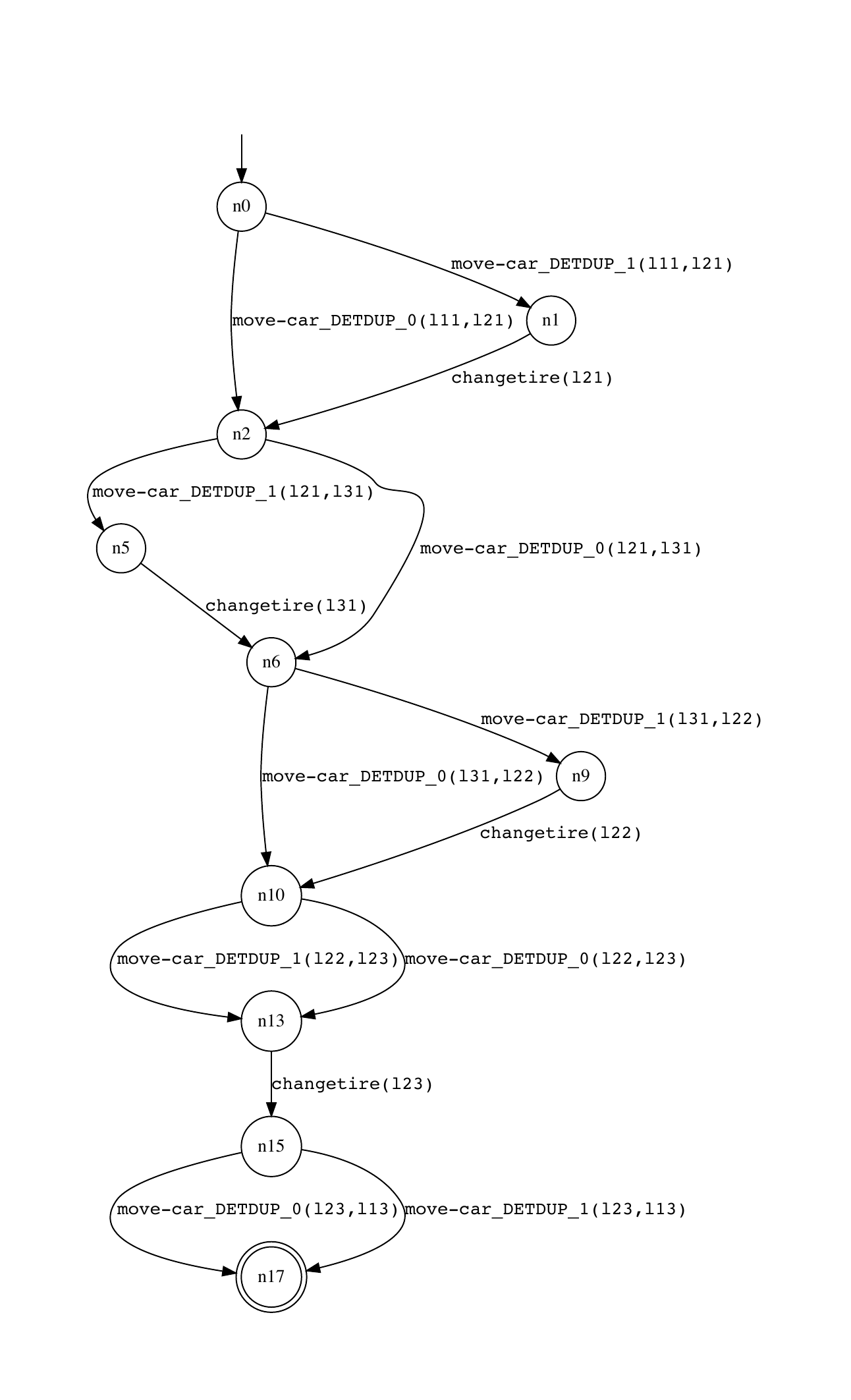}
\caption{The transition system showing policies found without the \texttt{trans} operator} 
\label{fig:policy-no-trans}
\end{figure}

A plan is whatever path from $n0$ leading to state $ng$. 
Moreover, as we can see from Figure \ref{fig:policy-trans}, there is a perfect alternation between domain's actions and the \texttt{trans} action. Additionally, the above-mentioned script could also remove transitions involving the \texttt{trans} action getting the final plan. We can see it in Figure \ref{fig:policy-no-trans}.

\section{Summary}
In this report, we have faced the problem of \FOND Planning for \LTLf/\PLTLf goals. In particular, we have proposed a new solution, called \FONDFOR, that essentially reduces the problem to a ``classical'' \FOND planning problem. This has been possible thanks to our \LTLfToDFA Python tool which has been employed for the encoding of temporally extended goals into standard \PDDL. 
Finally, we have seen examples of execution results of the \FONDFOR tool.

\clearpage
\bibliographystyle{plainnat}
\bibliography{main.bib}

\end{document}